\documentclass[runningheads]{llncs}
\usepackage{graphicx}
\usepackage{array, floatrow, booktabs}
\usepackage{tabularx}
\usepackage{amsmath}
\usepackage{amsfonts}
\usepackage{amssymb}
\usepackage{mathtools}
\usepackage{bm}
\usepackage{accents}
\usepackage[caption=false,font=footnotesize]{subfig}
\usepackage{microtype}
\usepackage[dvipsnames]{xcolor}
\usepackage{color, colortbl}
\definecolor{Gray}{gray}{0.9}
\newfloatcommand{capbtabbox}{table}[][\FBwidth]

\usepackage{hyperref}

\usepackage[capitalise,nameinlink]{cleveref}

\begin{document}
\title{An Explainable Geometric-Weighted Graph Attention Network for Identifying Functional Networks Associated with Gait Impairment}
\titlerunning{xGW-GAT: Explainable, Geometric-Weighted GAT}

\author{Favour Nerrise\inst{1} \and Qingyu Zhao\inst{2} \and Kathleen L. Poston\inst{3} \and Kilian M. Pohl\inst{2} \and Ehsan Adeli\thanks{Corresponding author.}\inst{2}} 

\authorrunning{F. Nerrise et al.}

\institute{Department of Electrical Engineering, Stanford University, Stanford, CA, USA \and 
Dept. of Psychiatry \& Behavioral Sciences, Stanford University, Stanford, CA, USA \and 
Dept. of Neurology \& Neurological Sciences, Stanford University, Stanford, CA, USA \\ 
\email{\{fnerrise,eadeli\}@stanford.edu}\\
}

\maketitle              
\begin{abstract}
One of the hallmark symptoms of Parkinson's Disease (PD) is the progressive loss of postural reflexes, which eventually leads to gait difficulties and balance problems. Identifying disruptions in brain function associated with gait impairment could be crucial in better understanding PD motor progression, thus advancing the development of more effective and personalized therapeutics. In this work, we present an explainable, geometric, weighted-graph attention neural network \textbf{(xGW-GAT)} to identify functional networks predictive of the progression of gait difficulties in individuals with PD. \textbf{xGW-GAT} predicts the multi-class gait impairment on the MDS-Unified PD Rating Scale (MDS-UPDRS). Our computational- and data-efficient model represents functional connectomes as symmetric positive definite (SPD) matrices on a Riemannian manifold to explicitly encode pairwise interactions of entire connectomes, based on which we learn an attention mask yielding individual- and group-level explainability. Applied to our resting-state functional MRI (rs-fMRI) dataset of individuals with PD, \textbf{xGW-GAT} identifies functional connectivity patterns associated with gait impairment in PD and offers interpretable explanations of functional subnetworks associated with motor impairment. Our model successfully outperforms several existing methods while simultaneously revealing clinically-relevant connectivity patterns. The source code is available at {\small \url{https://github.com/favour-nerrise/xGW-GAT}}.
\end{abstract}

\section{Introduction}
\label{introduction}
Parkinson’s Disease (PD) is an age-related neurodegenerative disease with complex symptomology that significantly impacts the quality of life, with nearly 90,000 people diagnosed each year in North America ~\cite{willis2022incidence}. Recent research has shown that gait difficulty and postural impairment symptoms of PD are highly correlated with alterations in various brain networks, including the motor, cerebellar, and cognitive control networks ~\cite{togo2023interactions}. Understanding brain functional networks associated with an individual's gait impairment severity is essential for developing targeted interventions, such as physical therapy or brain stimulation techniques. However, most prior works have \textit{either} focused only on a binary diagnosis (PD vs.~Control) \cite{li2018resting} (ignoring the progression and heterogeneity of the disease symptoms) \textit{or} only used sensor- and vision-based technologies ~\cite{lu_vision-based_2020,endo_gaitforemer_2022} to quantify PD symptoms (abstaining from identifying brain networks associated with gait impairment severity). 

Graph Neural Networks (GNNs) have been highly successful in inferring neural activity patterns in resting-state fMRI (rs-fMRI) \cite{rubinov2010complex}. These models represent functional connectivity matrices as weighted graphs, where each node is a brain region of interest (ROI), and the edges between them capture the magnitude of connectivity, i.e., interactions, as weights. Changes in the connectivity strengths can reflect intrinsic representations in a high-dimensional space that correlate with symptom or disease severity. Assuming that edges with higher weights exert greater functional connectivity (and vice versa), GNNs can encode how ROIs and their neighbors across various individuals can possess similar attributes. GAT \cite{velickovic2017graph} is a well-known GNN model that encodes pairwise interactions (edges) into an attention mechanism and uses eigenvectors and eigenvalues of each node as positional embeddings for local structures. However, since each node or ROI in a brain network has the same degree and connects to every other node, standard graph representations are limited in modeling functional connectivity differences in a high-dimensional space that can be used for inter-subject functional covariance comparison. Riemannian geometry \cite{klingenberg1959contributions} is another robust, mathematical framework for rs-fMRI analysis that projects a functional, connectivity matrix in a manifold of symmetric positive-definite (SPD) matrices, making it possible to model high-dimensional, edge interactions and dependencies. It has been applied to analyzing gait patterns ~\cite{olmos2023gait} in Parkinson's disease and to functional brain network analysis in other neurological disorders (e.g., Mild Cognitive Impairment \cite{dodero_kernel_2015} and autism \cite{wong2018riemannian}). 

Addressing the problem of identifying brain functional network alterations related to the severity of gait impairments presents several challenges: (\textbf{i}) clinical datasets are often sparse or highly imbalanced, especially for severely impaired disease states; (\textbf{ii}) although substantial progress has been made in modeling functional connectomes using graph theory, few studies exist that capture the individual variability in disease progression and they often fall short of generating clinically relevant explanations that are symptom-specific. 

In this work, we propose a novel, explainable, geometric weighted-graph attention network (\textbf{xGW-GAT}) that embeds functional connectomes in a learnable, graph structure that encodes discriminative edge attributes used for attention-based, transductive classification tasks. We train the model to predict a gait impairment rating score (MDS-UPDRS Part 3.10) for each PD participant. To mitigate  limited clinical data across all different classes of gait impairment and data imbalance (challenge \textbf{i}), we propose a stratified, learning-based sample selection method that leverages non-Euclidean, centrality features of connectomes to sub-select training samples with the highest predictive power. To provide clinical interpretability (challenge \textbf{ii}), \textbf{xGW-GAT} innovatively produces individual and global attention-based, explanation masks per gait category and soft assigns nodes to functional, resting-state brain networks. We apply the proposed framework on our dataset of 35 clinical participants and compare it with existing methods. We observe significant improvements in classification accuracy while enabling adequate clinical interpretability.

In summary, our contributions are:
(1) we propose a novel, geometric attention-based model, \textbf{xGW-GAT}, that uses edge-weights to depict neighborhood influence from local node embeddings during dynamic, attention-based learning;
(2) we develop a multi-classification pipeline that mitigates sparse and imbalanced sampling with stratified, learning-based sample selection during training on real-world clinical datasets; 
(3) we provide an explanation generator to interpret attention-based, edge explanations that highlight salient brain network interactions for gait impairment severity states;
(4) we establish a new benchmark for PD gait impairment assessment using brain functional connectivities.

\section{xGW-GAT: Explainable, Geometric-Weighted GAT}\vspace{-10pt}
\label{methods}
\paragraph{Problem definition.} Assume a set of functional connectomes, ${\mathcal{G}_n \in \mathbb{R}^{d\times d}, \mathcal{G}_2, \ldots, \mathcal{G}_N}$ are given, where $N$ is the number of samples and $d$ is the number of ROIs. Each connectome is represented by a weighted, undirected graph $\mathcal{G} = (\mathcal{V},\mathcal{E},\mathbf{W})$, where $\mathcal{V} = {\{v_i\} }^{d}_{i=1}$ is the set of nodes, $\mathcal{E} \subseteq \mathcal{V} \times \mathcal{V}$ is the edge set, and $\mathbf{W} \in \mathbb{R}^{|\mathcal{V}| \times |\mathcal{V}|}$ denotes the matrix of edge weights. The weight $w_{ij}$ of an edge $e_{ij} \in \mathcal{E}$ represents the strength of the functional connection between nodes $v_i$ and $v_j$, i.e., the Pearson correlation coefficient of the time series of the pair of the nodes. Each $\mathcal{G}_{n}$ contains node attributes $X_{n}$ and edge attributes $H_{n}$. We  develop a model that predicts a gait impairment score, $\mathcal{Y}_{n}$ and outputs an individual explanation mask $\mathbf{M}_{c} \in \mathbb{R}^{d \times d}$ per class $c$ to assign ROIs to functional brain networks. 

\begin{figure}[t]
\includegraphics[width=1\linewidth]{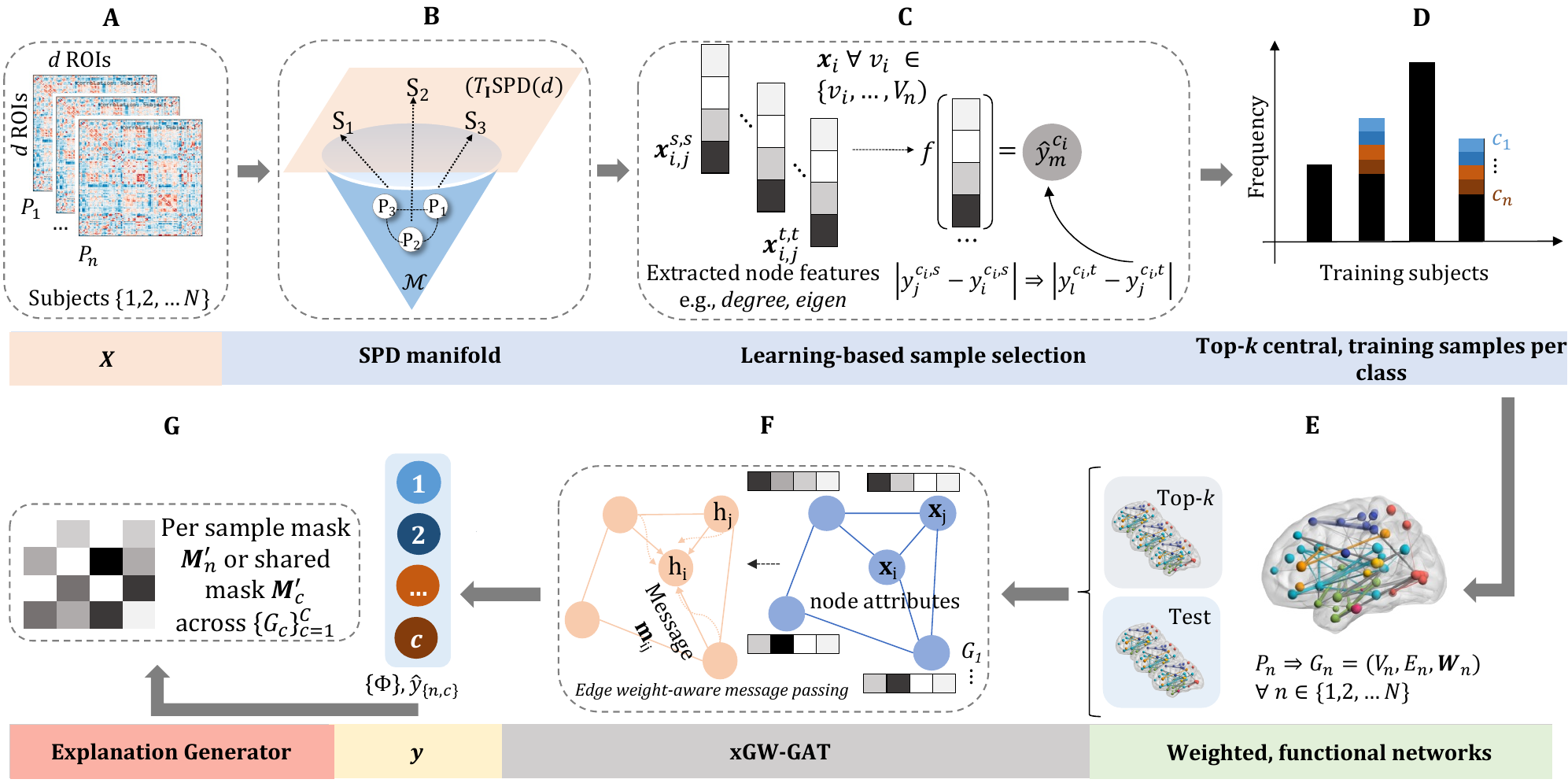}
\caption{\textbf{xGW-GAT}. (A) Input: functional connectomes. (B) Extract pairwise tangent matrices in SPD($d$). (C) Compress tangent matrices into weighted graphs (connectomes) (D) Use linear regression to train a mapping, $f$, on training samples to learn pairwise differences between target and record scores. (E) Group top-$k$ samples per class across $N$-fold cross-validation runs with the lowest predicted difference and oversample for imbalance. (F) Represent samples as weighted, graphs and use edge weight-aware attention to encode and propagate learning; predict gait score. (G) Produce explanation masks for each class or individual participants within functional brain networks.}
\label{fig:xGW-GAT-pipeline}
\end{figure}
\subsection{Connectomes in a Riemmanian Manifold} 
Functional connectivity matrices belong to the manifold of symmetric positive-definite (SPD) matrices \cite{you2021re}. We leverage Riemmanian geometry to perform principled comparisons between different connectomes, such as prior work \cite{shahbazi2021using}. To highlight connections between adjacent nodes, each weight matrix $\mathbf{W}_n \in \mathbb{R}^{d\times d}$ can be represented as a symmetric, adjacency matrix with zero, non-negative eigenvalues, where each element of the adjacency matrix is the edge weight, $w_{ij}$ between nodes $i$ and $j$. We then consider $\mathbf{W}_n$ to be a point, $\mathbf{S}_n$, in the manifold of SPD matrices $Sym_d^{+}$ that locally looks like a topological Euclidean space. However, $Sym_d^{+}$ does not form a vector space; thus, we project each SPD matrix $\mathbf{S}_n$ onto a common tangent space using parallel transport. Given a reference point $\mathbf{S}_i \in Sym_d^{+}$, we transport a tangent vector $v \in T_{\mathbf{S}_j}$ from $\mathbf{S}_j$ to $\mathbf{S}_i$ along the geodesic connecting $\mathbf{S}_j$ and $\mathbf{S}_i$ (see \cref{fig:xGW-GAT-pipeline}-B). This process is performed for each subject $n = 1, 2, \ldots, N$, yielding a set of tangent vectors in a common tangent space that can be analyzed using traditional Euclidean methods.

 To calculate the geodesic distance between two SPD matrices $\mathbf{S}_i$ and $\mathbf{S}_j \in Sym_d^{+}$, we adopt the Log-Euclidean Riemannian Metric (LERM) \cite{arsigny2007geometric}, $\mathcal{D}_{le}$ as follows:  
    \begin{equation} 
        \label{eq:loge}
        \mathcal{D}_{le}(\mathbf{S}_i, \mathbf{S}_j) = \|log(\mathbf{S}_i) - log(\mathbf{S}_j)\|^{2}_F
    \end{equation}
where $\| \cdot \|_F$ is the Frobenius norm. LERM is invariant to similarity transformations (scaling and orthogonality) and is computationally efficient for high-dimensional data. See the Supplementary Material for results with other distance metrics.

\subsection{Stratified Learning-based Sample Selection} 
Data availability and dataset imbalance are re-occurring challenges with real-world clinical datasets, often leading to bias and overfitting during model training. We address this by expanding a learning-based sample selection method \cite{hanik_predicting_2022} to weight per-class distributions. We assume that similar brain connectivity networks are correlated with disease severity whereas  connectomes that vary in topological patterns might elicit different gait impairment scores. Our sub-sampling technique selects training samples containing the highest representative power, i.e., contributing the least amount of pairwise differences for predicting a gait score. We divide our training samples into subgroups: train-in, $n_s$, and holdout, $n_t$ using $N$-fold cross-validation. For each pair of symmetric $d$-by-$d$ tangent matrices, $\mathbf{S}^{s,s}_{i,j} \in T_\mathbf{I} \text{SPD}(d)$, we encode the pairwise differences between the connectomes from the train-in, $n_s$, to obtain a set of $n_s (n_s - 1)/2$ tangent matrices in $T_{\mathbf{I}} \text{SPD}(d)$. Each tangent matrix represents the “difference” between two connectomes. We affix a threshold of $k$ samples to be selected from each class $c$ to identify $l$ central training samples with the highest expected predictive power, i.e., the \emph{lowest average difference} in target gait impairment scores per class, $\hat{y}_c$, between samples $j$ from the train-in and holdout group. We select degree, closeness, and eigenvector centrality as our topological features that encode information on \emph{changes} in node connectivity.  We train a linear regression mapping $f$ on the Riemannian geometric distances $\mathcal{D}_{le}(\mathbf{S}^s_i, \mathbf{S}^s_j)$ between the connectomes from $n_s$ using the vectorized upper triangular portion (including the diagonal) of the tangent matrices. The \emph{absolute difference in target score, per class}, between samples $i$ and $j$ from the train-in group $n_s$ is denoted by $|\hat{y}^s_{c,j} - \hat{y}^s_{c,i}|$ (see \cref{fig:xGW-GAT-pipeline}-C). The top-$k$ samples per class with the highest predictive power are sub-selected from the total training set, oversampled for class imbalance with RandomOverSampler~\cite{gui2017imb}, and used for training xGW-GAT layers (see \cref{fig:xGW-GAT-pipeline}-D).

\subsection{Dynamic Graph Attention Layers}{\label{xGW-GAT}}
\noindent\textbf{Attention.} We employ Graph Attention Network version 2 (GATv2) ~\cite{brody2021attentive}, a GAT \cite{velickovic2017graph} variant to perform dynamic, multi-head, edge-weight attention message passing for classifying each $\mathbf{S}_n$. We assume that every node $i \in \mathcal{V}$ has an initial representation $\mathbf{h}_{i}^{(0)} \in \mathbb{R}^{d_0}$. GATv2 updates each node representation, $\mathbf{h}$ based on the features of neighboring nodes and the edge weights between nodes by computing attention scores $\alpha_{ij}$ for every edge $(i,j)$ by normalizing attention coefficients $e(\mathbf{h}_{i}, \mathbf{h}_{j})$. $\alpha_{ij}$ measures the importance of node $j$'s features to node $i$'s at layer $l$ by performing a weighted sum over the neighboring nodes $j \in \mathcal{N}_i$: 
\begin{equation}
  e(\mathbf{h}_{i}, \mathbf{h}_{j})\coloneqq \text{LeakyReLU}(\mathbf{a}^{(l)\top}[\mathcal{\mathbf{\Theta}}^{(l)}\mathbf{h}^{(l-1)}_i \| \mathcal{\mathbf{\Theta}}^{(l)}\mathbf{h}^{(l-1)}_j])
\end{equation}
\begin{equation} \label{eq:alpha}
    \alpha_{ij} \coloneqq softmax_{j}(e(\mathbf{h}_{i}, \mathbf{h}_{j}))
\end{equation}
\begin{equation}
    \mathbf{h}^{(l+1)}_i \coloneqq \sigma \left(\sum\limits_{j \in \mathcal{N}_i} \alpha^{(l)}_{ij} \mathbf{h}^{(l-1)}_j\right).
\end{equation}
where $\mathbf{a}^{(l)} \in \mathbb{R}^{2F}$ and  $\mathcal{\mathbf{\Theta}}^{(l)}$ are trainable parameters and learned,  $\mathbf{h}^{(l)}_i \in \mathbb{R}^F$ is the embedding for node $i$, $\sigma$ represents a non-linearity activation function, and $\|$ denotes vector concatenation. 
As conventional graph attention mechanisms for transductive tasks typically do \textit{not} incorporate  edge attributes, we introduce an attention-based, message-passing mechanism incorporating edge weights, similar to \cite{cui2022interpretable}. The algorithm uses a message vector $\mathbf{m}_{ij} \in \mathbb{R}^{F}$ by concatenating node features of neighboring nodes $i, j$, and edge weight $\mathbf{W}_{i,j}$:
\begin{equation}
    \mathbf{m}^{(l)}_{ij} = \text{MLP}_1([\mathbf{h}^{(l)}_i; \mathbf{h}^{(l)}_j; \mathbf{W}_{ij}]),
\end{equation}
where $\text{MLP}_1$ is a Multi-Layer Perceptron. Accordingly, an update of each ROI representation is influenced by its neighboring regions weighted by their connectivity strength. After stacking $L$ layers, a readout function summarizing all node embeddings is employed to obtain a graph-level embedding $\mathbf{g}$: 
\begin{equation}
    \mathbf{z} = \sum\limits_{i \in V} \mathbf{h}^{(L)}_i, \mathbf{g} = \text{MLP}_2(\mathbf{z}) + \mathbf{z}.
\end{equation}

\vspace{-8pt}\noindent\textbf{Loss function.} \textbf{xGW-GAT} layers (\cref{fig:xGW-GAT-pipeline}-F) are trained with a supervised, weighted negative log-likelihood loss function to mitigate class imbalance across classes, $C$, defined as:
\begin{equation}
     \mathcal{L}_\text{NLL} \coloneqq \frac{-1}{N} \sum\limits^N_{p=1} \sum\limits^C_{q=1} r_q y_{pq} log(\hat{y}_{pq}),
\end{equation}
where $r_q$ is the rescaling weight for the $q$-th class, $y_{pq}$ is the $q$-th element of the true label vector $y_p$ for the $p$-th sample, and $\hat{y}_{pq}$ is the predicted label vector.

\subsection{Individual- and Global-Level Explanations.} 
We define an attention explanation mask for each sample, $n \in {1,2,\ldots,N}$ and for each class $c \in {1, 2, \ldots, C}$ that identifies the most important node/ROI connections contributing to the classification of subjects. We return a set of attention coefficients $\mathbf{\alpha}^{n} = [\alpha_1^n, \alpha_2^n, \ldots, \alpha_S^n]$ for each sample $n$, where $S$ is the number of attention heads. We aggregate trained, attention coefficients per sample used for predicting each $\hat{y}$ using a $max$ operation that returns $\mathbf{\alpha}^{n}_{\text{max}} \in \mathbb{R}^{d \times d}$. An explanation mask per class, $M_c$, or per sample, $M_n$, can be derived using the max attention coefficients, $\mathbf{\alpha}_{\text{max}}$ (\cref{fig:xGW-GAT-pipeline}-G):
\begin{equation} \label{eq:mask}
    \mathbf{M}_c = \frac{1}{N} \sum_{n=1}^{N} \mathbf{\alpha}_{\text{max}}; \quad
    \mathbf{M}_n = \frac{1}{C} \sum_{c=1}^{C} \mathbf{\alpha}_{\text{max}}.
\end{equation}
$\mathbf{M}$ can be soft-thresholded to retain the top-$L$ most positively attributed attention weights to the mask as follows:
\begin{equation} \label{eq:soft_thresh}
    \mathbf{M}^{'}[i] = 
    \begin{cases} 
    \mathbf{M}[i] & \text{if } \mathbf{M}[i] \in \text{Top-L}(\mathbf{M}) \\
    0, & \text{otherwise}, 
    \end{cases}
\end{equation}  
where $\text{Top-L}(\mathbf{M})$ represents the set of top-$L$ elements in $\mathbf{M}$. 

\section{Experiments}

\noindent\textbf{Dataset.} 
We obtained data from a private dataset ($n=35$, mean age 69$\pm 7.9$) defined in ~\cite{lu2021quantifying}, which contains MDS-UPDRS exams from all participants. Following previously published protocols ~\cite{poston2016compensatory}, all participants are recorded during the off-medication state. Participants were evaluated by a board-certified movement disorders specialist on a scale from 0 to 4 based on MDS-UPDRS Section 3.10 ~\cite{goetz2019mds}. The dataset includes 22 participants with a score 1, 4 participants with a score 2, 4 participants with a score of 3, 4 participants with a score 4, and 1 participant with a score 0 on MDS-UPDRS item 3.10. The single score-0 participant (normal) was combined with the score-1 participants (minor gait impairment) to adjust for severe class imbalance. We pre-processed functional connectivity matrices and corrected them for possible motion artifacts using the CONN toolbox \cite{whitfield2012conn}. The FC matrices were obtained using a combined Harvard-Oxford and AAL parcellation atlas \cite{whitfield2012conn} with 165 ROIs, where each entry in row $i$ and column $j$ in the matrix is the Pearson correlation between the average rs-fMRI signal measured in ROI $i$ and ROI $j$. We imputed any missing ROI network scores with the mean score per column and Z-transformed FC matrices $\left[ \mu=0,\sigma=1 \right]$. This dataset (like other clinical datasets in practice) poses highly imbalanced distributions for classes with severe impairment, which makes it useful to demonstrate our method's capability in an imbalanced and limited-data scenario. In addition, most existing studies focus on differentiating participants from controls, while the severity of specific impairments is understudied (our focus).

\noindent\textbf{Software.} All experiments were implemented in Python 3.10 and ran on Nvidia A100 GPU runtimes. We used PyTorch Geometric~\cite{fey2019fast}, PyTorch, and Scikit-learn for machine learning methods. We used the SPD class from the Morphometrics package to compute Riemannian geometrics and NetworkX to extract the topological features of the graphs from the tangent matrices. Hyper-parameters are tuned automatically with the open-source AutoML toolkit NNI ({\small \url{https://github.com/microsoft/nni})}. 

\noindent\textbf{Setup.} We used the mean, connectivity profile, i.e., $\mathbf{W}$ \cite{cui2022braingb}, as the node feature for xGW-GAT layers, a weighted ADAM optimizer, a learning rate of $1e-4$, a batch size of $2$ for training and $1$ for test, and $100$ training epochs. We used $2$ GATv2 layers, dropout rate=$0.1$, hidden\_dim=$8$,  heads$=2$, and a global mean pooling layer. We used $4$-fold cross-validation to partition training and holdout sets and selected $k=4$ as the optimal number of selected training samples between $2$ and $15$. We report weighted, macro average scores for $F_1$, area under the ROC curve (AUC), precision (Pre), and recall (Rec) over $100$ trials.

\subsection{Results}
We perform a multi-class classification task of \emph{Slight(1), Mild(2), Moderate(3), Severe(4)} gait impairment severity. To benchmark our method, we compare our results with several state-of-the-art classifiers: GAT \cite{velickovic2017graph}, GCN \cite{kipf2016semi}, PNA \cite{corso2020principal}, and two state-of-the-art deep models design for brain networks: BrainNetCNN \cite{kawahara2017convolutional} and BrainGNN \cite{li2021braingnn}. We also perform an ablation study on sample selection and the type of topological features used in training our method: (!ss) no [stratified, learning-based] sample selection, (dc) node degree centrality, (cc) node closeness centrality, and (ec) eigenvector centrality. Results for the highest-performing settings of \textbf{xGW-GAT} are displayed in \cref{table:comparison_performance} and node feature descriptions are included in the Supplementary Material.

\begin{figure}[t]
\begin{floatrow}
\capbtabbox{%
\resizebox{.47\textwidth}{!}{
    \begin{tabular}{ lcccc } 
        \toprule
        \textbf{Method} & \textbf{Pre} & \textbf{Rec} & $\mathbf{F_{1}}$ & \textbf{AUC} \\
        \midrule
        GCN* \cite{kipf2016semi}& 0.46 & 0.48 & 0.47 & 0.54\\
        PNA*  \cite{corso2020principal} & 0.52 & 0.54 & 0.53 & 0.56\\
        BrainNetCNN*\cite{kawahara2017convolutional}& 0.62 & 0.71 & 0.66 & 0.57\\
        BrainGNN*\cite{li2021braingnn}& 0.66 & 0.53 & 0.59 & 0.62\\
        GAT* \cite{velickovic2017graph}& 0.70 & 0.58 & 0.64 & 0.71\\
        \midrule
        \textbf{xGW-GAT} (dc)* & 0.61 & 0.65 & 0.63 & 0.51 \\
        \textbf{xGW-GAT} (ec)* & 0.64 & 0.62 & 0.63 & 0.72\\
        \textbf{xGW-GAT} (cc)* & 0.61 & 0.53 & 0.57 & 0.57\\
        \textbf{xGW-GAT} (!ss)* & 0.55 & 0.47 & 0.51 & 0.54\\
        \midrule 
        \rowcolor{Gray}
        \textbf{xGW-GAT} (ss) & \textbf{0.75} & \textbf{0.77} & \textbf{0.76} & \textbf{0.83}\\
        \bottomrule
    \end{tabular}
}
}
{%
  \caption{Comparison with baseline and ablated methods. * indicates statistical difference by Wilcoxon signed rank test at ($p < 0.05$) compared with our method.}%
  \label{table:comparison_performance}
}
\ffigbox{%
  \includegraphics[width=0.85\linewidth]{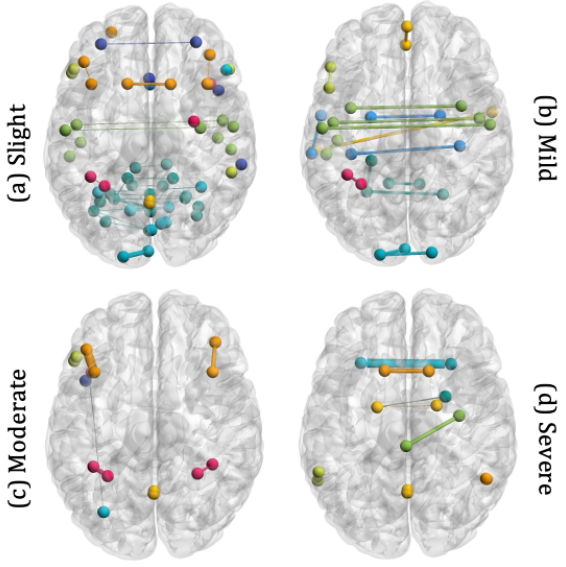}\vspace{-10pt}
}{%
  \caption{Salient ROI connections on explanation brain networks across the four classes of gait impairment (\textcolor{Dandelion}{DMN}, \textcolor{LimeGreen}{SMN}, \textcolor{Turquoise}{VN}, \textcolor{BlueViolet}{SN}, \textcolor{WildStrawberry}{DAN}, \textcolor{BurntOrange}{FPN}, \textcolor{SpringGreen}{LN}, \textcolor{PineGreen}{CN}, and \textcolor{Cerulean}{BLN}).}%
  \label{fig:glass_brains}
}
\end{floatrow}
\end{figure}

The results (\cref{table:comparison_performance}) show that \textbf{xGW-GAT} yields significant improvement in performance over SOTA graph-based models, including models designed for brain network analysis. xGW-GAT with our stratified, learning-based selection method combined with the RandomOverSampler technique to temper the effects of class imbalance outperforms a standard xGW-GAT by 42\%. Compared with SOTA deep models like GCN and PNA, our model also outperforms them by large margins, with up to 29\% improvement for AUC. These predictions are promising for an explainable analysis of PD gait impairment while also minimizing random uncertainties introduced in individual participant graphs. 

\section{Discussion}
\emph{Brain Networks Mapping.} As shown in \cref{fig:glass_brains}, we aid interpretability for clinical relevance by partitioning the ROIs into nine ``networks" based on their functional roles: Default Mode Network \textcolor{Dandelion}{(DMN)}, SensoriMotor Network \textcolor{LimeGreen}{(SMN)}, Visual Network \textcolor{Turquoise}{(VN)}, Salience Network \textcolor{BlueViolet}{(SN)}, Dorsal Attention Network \textcolor{WildStrawberry}{(DAN)}, FrontoParietal Network \textcolor{BurntOrange}{(FPN)}, Language Network \textcolor{SpringGreen}{(LN)}, Cerebellar Network \textcolor{PineGreen}{(CN)}, and Bilateral Limbic Network \textcolor{Cerulean}{(BLN)} are colored accordingly, while edges across different systems are colored gray. Edge widths here are the attention weights.

\noindent\emph{Salient ROIs.} We provide per-class and individual-level interpretations for understanding how ROIs contribute to predicting gait impairment scores. We build the node and edge files with the thresholded attention explanation masks, $\mathbb{M^{'}}$ per PD participant or per class and plot glass brains using BrainNet Viewer ({\small \url{(https://www.nitrc.org/projects/bnv/)}).

We observe that rich interactions decrease significantly for the \emph{Mild} class, \cref{fig:glass_brains}(b), within the \textcolor{PineGreen}{CN}, primarily associated with coordinating voluntary movement, the \textcolor{BlueViolet}{SN}, responsible for thought, cognition, and planning behavior, and the \textcolor{Turquoise}{VN}, the center for visual processing during resting and task states. These observations are consistent with existing neuroimaging findings, which support that PD is positively associated with the severity of cognitive deficits and neuromotor control for inter-network and intra-network interactions within the salience network, cerebellar lobules, and visual network \cite{zhu2019abnormal,ruan2020impaired}. Similarly, there are significantly lower connections within CN and VN and sparser connections within the \textcolor{LimeGreen}{SMN} for the \emph{Moderate} and \emph{Severe} classes, \cref{fig:glass_brains}(c)-(d). Existing studies show functional connectivity losses within the sensorimotor network (SMN)  \cite{caspers2021within} are correlated with  disruptions in regional and global topological organization for SMN areas for people with PD, resulting in loss of motor control. For \emph{Mild}, \emph{Moderate}, and \emph{Severe} PD participants, abrupt connectivity is also  observed for the frontoparietal network, \textcolor{BurntOrange}{FPN}, known for coordinating behavior and associated with connectivity alterations correlated with motor deterioration \cite{vervoort2016functional}. 

\section{Conclusion}
This study showcases a novel benchmark for using an explainable, geometric-weighted graph attention network to discover patterns associated with gait impairment. The framework innovatively integrates edge-weighted attention encoding and explanations to represent neighborhood interactions in functional brain connectomes, providing interpretable functional network clustering for neurological analysis. Despite a small sample size and imbalanced settings, the lightweight model offers stable results for quick inference on categorical PD neuromotor states. Future work includes new experiments, an expanded, multi-modal dataset, and sensitivity and specificity analysis to discover subtypes associated with the severity of PD gait impairment. 
\\
\\
{\small \noindent\textbf{Acknowledgements.} This work was partially supported by NIH grants  (AA010723,
NS115114, P30AG066515), Stanford School of Medicine Department of Psychiatry and Behavioral Sciences Jaswa Innovator Award, UST (a Stanford AI Lab alliance member), and the Stanford Institute for Human-Centered AI (HAI) Google Cloud credits.} FN is funded by the Stanford Graduate Fellowship and the Stanford NeuroTech Training Program Fellowship. 
\bibliographystyle{splncs04}
\bibliography{main}
\clearpage
\appendix

\title{Supplementary Material for ``An Explainable Geometric-Weighted Graph Attention Network for Identifying Functional Networks Associated with Gait Impairment"}
 \titlerunning{Supplementary Material for xGW-GAT}
\author{}
\institute{}
\maketitle    
\vspace{-10ex}
\section*{Riemmanian Metrics}
The Riemannian metric is a geometric structure that assigns a positive-definite inner product to each tangent space of a smooth manifold. This allows us to define notions of length, angle, and curvature on the manifold. Let $\mathcal{M}$ be a smooth manifold, the Riemannian metric on $\mathcal{M}$ is a smoothly varying family of inner products $g_p$ on the tangent spaces $\mathbf{T}_p\mathcal{M}$, for each point $p \in \mathcal{M}$.
\\
\noindent \textbf{Affine Invariant Riemannian Metric (AIRM)}: AIRM\footnote{Pennec, X., Fillard, P. and Ayache, N.: A Riemannian framework for tensor computing. International Journal of computer vision, 66, pp.41-66. (2006)} is an intrinsic Riemannian metric that characterizes the geometry of a symmetric positive definite (SPD) space by returning a number signifying the geodesic distance between two elements in tangent space. Given two SPD matrices $\mathbf{P}$ and $\mathbf{Q}$, the AIRM distance is defined as:
\begin{equation*}
    d_{\text{AIRM}}(\mathbf{P}, \mathbf{Q}) = \sqrt{\sum_{i=1}^{n} \log^2 \lambda_i}
\end{equation*}
where $\lambda_i$ are the eigenvalues of $\mathbf{P}^{-1}\mathbf{Q}$. AIRM is invariant to affine transformations, which makes it robust in many applications, and is strictly bound by innate geometric conditions.
\\
\noindent \textbf{Log-Euclidean Riemannian Metric (LERM)}: LERM\footnote{Arsigny, V., Fillard, P., Pennec, X., Ayache, N.: Geometric means in a novel vector space structure on symmetric positive-definite matrices. SIAM journal on matrix analysis and applications 29(1), 328–347 (2007)} is an extrinsic, closed-form formula for the Riemannian mean and is used to measure the geometric difference between matrices in a Riemannian manifold. Given two SPD matrices $\mathbf{P}$ and $\mathbf{Q}$, the LERM distance is defined as:
\begin{equation*}
  d_{\text{LERM}}(\mathbf{P}, \mathbf{Q} = \|log(\mathbf{P}) - log(\mathbf{Q})\|^{2}_F
\end{equation*}
where $|\cdot|^{2}_F$ denotes the Frobenius norm. While not affine-invariant, LERM is particularly advantageous because it embeds points in SPD matrices into a Euclidean space, making them amenable to standard Euclidean distance computations while preserving geometric information and thus is more computationally efficient than affine-invariant metrics like AIRM. 
\\
\noindent \textbf{\emph{symmetrized} Kullback-Leibler Divergence Metric (sKLDM)}: sKLDM\footnote{Kullback, S. and Leibler, R.A.: On information and sufficiency. The annals of mathematical statistics, 22(1), pp.79-86. (1951)} is an extension of the asymmetric KL divergence metric (KLDM), which quantifies the dissimilarity between probability distributions on a Riemannian manifold while taking into account the symmetrical nature of their comparison. It is also known as Jeffreys divergence. Given two SPD matrices $\mathbf{P}$ and $\mathbf{Q}$, the KLDM geodesic distance is defined as:
\begin{equation*}
    d_{\text{KLDM}}(\mathbf{P}, \mathbf{Q}) = Tr(\mathbf{P}(log(\mathbf{P})- log(\mathbf{Q}))
\end{equation*}
where $Tr(\cdot)$ denotes the trace operator. The symmetrized function is as follows:
\begin{equation*}
  d_{\text{sKLDM}}(\mathbf{P}, \mathbf{Q}) = \frac{1}{2} \left( d_{\text{KLDM}}(\mathbf{P}, M) + d_{\text{KLDM}}(\mathbf{Q}, M) \right)
\end{equation*}
where $M$ is the average of the two distributions, i.e. $M = \frac{1}{2}(\mathbf{P}+\mathbf{Q})$.

\begin{table}[t]
\caption{Ablation study for Riemannian metrics: *Note that the LERM results are the same as our best, reported results in the main paper.}
\label{app_dmetrics}
 \begin{tabular}{ lcccc } 
        \toprule
        \textbf{Method} & \textbf{Pre} & \textbf{Rec} & $\mathbf{F_{1}}$ & \textbf{AUC} \\
        \midrule
        \textbf{xGW-GAT} (AIRM) & 0.70 & 0.67 & 0.61 & 0.67 \\
        \textbf{xGW-GAT} (sKLDM) & 0.55 & 0.45 & 0.49 & 0.51 \\
        \midrule 
        \rowcolor{Gray}
        \textbf{xGW-GAT} (LERM)* & \textbf{0.75} & \textbf{0.77} & \textbf{0.76} & \textbf{0.83}\\
        \bottomrule
    \end{tabular}
\end{table}

\section*{Sample Selection Node Centrality Measures}
For each training sample, we derive the following node features that encode node centrality measures, i.e., a node's importance or influence within the network, based on criteria such as the number, quality, and proximity of its connections.
\begin{itemize}
     \item \emph{Degree Centrality}: Degree centrality of a node $j$, denoted $C_D(j)$, is the count of its direct connections or edges, defined as: 
     \begin{equation*}C_D(j) = \text{deg}(j)
     \end{equation*}
     where $\text{deg}(j)$ is the degree of node $d$.
    \item \emph{Eigenvector Centrality}: Eigenvector centrality of a node $j$, denoted $C_E(j)$, measures its influence based on the quality of its connections, defined as:
    \begin{equation*}
        C_E(j) = \frac{1}{\lambda} \sum_{i \in N(j)} A_{ij} C_E(i)
    \end{equation*}
    where $A_{ij}$ is the adjacency matrix, $N(j)$ is the set of neighbors of $i$, and $\lambda$ is the largest eigenvalue.
    \item \emph{Closeness Centrality}: Closeness centrality of a node $v$, denoted $C_C(j)$, measures the inverse average shortest path length to all other nodes, defined as:
    \begin{equation*}
        C_C(j) = \frac{1}{\sum_{u \neq v} d(j,i)}
    \end{equation*}
    where $d(j,i)$ is the shortest path length from $j$ to $i$.
\end{itemize}

\end{document}